\documentclass{article}

    \usepackage[preprint]{neurips_2025}

\usepackage[utf8]{inputenc} 
\usepackage[T1]{fontenc}    
\usepackage{hyperref}       
\usepackage{url}            
\usepackage{booktabs}       
\usepackage{amsfonts}       
\usepackage{nicefrac}       
\usepackage{microtype}      
\usepackage{xcolor}         
\usepackage{multirow}
\usepackage[pdftex]{graphicx}
\usepackage{amsmath}

\title{The Energy Cost of Reasoning: Analyzing Energy Usage in LLMs with Test-time Compute}

\author{%
  Yunho Jin \\
  Harvard University\\
  Cambridge, MA \\
  \texttt{yjin@g.harvard.edu} \\
  \And
  Gu-Yeon Wei \\
  Harvard University \\
  Cambridge, MA \\
  \AND
  David Brooks \\
  Harvard University \\
  Cambridge, MA \\
}

\begin{document}

\maketitle

\begin{abstract}
Scaling large language models (LLMs) has driven significant advancements, yet it faces diminishing returns and escalating energy demands. This work explores how test-time compute (TTC) can serve as an energy-efficient complement to conventional scaling strategies by allocating additional computational resources at inference time rather than during training. Specifically, we investigate whether employing TTC can achieve superior accuracy-energy trade-offs compared to simply increasing model size. Our empirical analysis reveals that TTC surpasses traditional model scaling in accuracy/energy efficiency, with notable gains in tasks demanding complex reasoning rather than mere factual recall. Further, we identify a critical interaction between TTC performance and output sequence length, demonstrating that strategically adjusting compute resources at inference time according to query complexity can substantially enhance efficiency. Our findings advocate for TTC as a promising direction, enabling more sustainable, accurate, and adaptable deployment of future language models.
\end{abstract}

\section{Introduction}
\label{sec:intro}

Test-time compute (TTC) has emerged as a promising approach to significantly enhance the performance of large language models (LLMs) without the reliance on traditional scaling laws. Conventionally, improving neural language models involved scaling either the number of parameters or expanding the training dataset~\citep{2020arXiv200108361K}. However, this approach demands substantial computational resources, resulting in considerable energy consumption during both training and inference phases. Furthermore, such scaling is increasingly constrained by the limited availability of sufficiently large and diverse datasets\citep{2022arXiv221104325V}, highlighting diminishing returns in terms of resource efficiency.

Recent research addresses this limitation by proposing a new strategy to enhance the performance of LLMs: allocating additional computational resources at inference time, known as TTC. TTC is inspired by human reasoning process, where humans reflect and reason logically to answer challenging questions rather than responding impulsively. Similarly, LLMs should leverage additional computational effort during inference to improve their responses. Prominent examples of TTC methodologies include prompting techniques like Chain-of-Thought (CoT) reasoning ~\citep{2022arXiv220311171W, 2023arXiv230510601Y}, self-revision methods~\citep{2024arXiv240601297K, 2024arXiv240615673L, 2023arXiv230317651M}, post-training for reasoning~\citep{deepseek, star, rest}, and ensemble-like approaches aggregating response from multiple parallel generations to improve the final response~\citep{llmonk, getting_50, 2023arXiv230306135I, 2022arXiv220311171W}. Despite methodological differences, these strategies collectively emphasize additional computational investment during inference to enhance model accuracy.

The rising computational demands of TTC necessitate careful consideration of associated energy costs. Recent industry reports underscore this concern, with Meta attributing up to 70\% of its AI power consumption to inference processes~\citep{wu2022sustainable}, Google reporting 60\% of machine learning energy usage~\citep{patterson2022carbon}, and AWS indicating inference-related demands accounting for 80-90\% of computing resources~\citep{barr2019amazon} all without TTC. Moreover, broader data center energy consumption trends indicate a rapid escalation, expected to constitute up to 12\% of total U.S. electricity usage by 2028~\citep{shehabi2024united}. The environmental impact and financial implications of such intensive energy consumption underline the urgency of evaluating the sustainability and efficiency of TTC approaches.

In this paper, we examine the accuracy-energy trade-offs associated with TTC compared to traditional model scaling. Specifically, we investigate the relationship between accuracy improvements and energy consumption incurred by employing TTC methods versus scaling model sizes across multiple benchmarks. Through comprehensive experiments conducted on Qwen2.5~\citep{qwen2.5} models (1.5B, 7B, 14B, and 32B parameters) using A100 GPUs across mathematical, coding, and common-sense reasoning benchmarks, we derive key insights into the comparative advantages and costs of TTC.

Our findings highlight several crucial observations:

\begin{itemize}
    \item TTC demonstrates superior accuracy-energy trade-off improvements over traditional model scaling. Notably, smaller models enhanced with TTC can outperform substantially larger models relying solely on scale.
    \item However, indiscriminate application of TTC can drastically elevate energy consumption, with cases where TTC usage amplifies energy costs up to 113.48$\times$, as observed with the 7B model on a coding benchmark.
    \item Output sequence length emerges as a reliable indicator of model comprehension, with extended sequences often signaling models' struggles or attempts at multiple reasoning pathways, consequently incurring higher energy usage.
    \item Complex questions drive models with TTC to allocate greater computational resources during inference, paralleling human cognitive processes in handling challenging tasks.
\end{itemize}

To the best of our knowledge, this study represents one of the first systematic explorations into the energy implications of TTC for LLM inference. Our analysis provides a foundational understanding of how TTC can serve as a more energy-efficient complement to traditional scaling. We explicitly focus on inference-related impacts, leaving exploration of training-phase considerations as important avenues for future research.
\section{Background}
\label{sec:background}

\paragraph{LLM Inference Pipeline}
Inference for transformer-based LLMs~\citep{llama, qwen2.5, gemini, gpt4, vaswani2017attention} typically involves two sequential stages: prefill and decode. Initially, when an input sequence is provided, it undergoes the prefill stage, where all tokens in the seqeunce are processed simultaneously. This stage computes the first output token and generates a Key-Value cache (KV-cache), which stores contextual information from the input sequence. Following the prefill stage, the model enters the decode stage, wherein it autoregressively generates subsequent tokens one-by-one, leveraging the previous KV-cache.

Each of these stages presents distinct computational bottlenecks. The prefill stage is predominantly compute-bound due to large matrix multiplications resulting from processing all tokens simultaneously. In contrast, the decode stage typically becomes memory bandwidth (BW)-bound because of the frequent accesses to the KV-cache.

\paragraph{Test-time Compute} Recently, leveraging additional computational resources at inference time—known as TTC—has emerged as an effective strategy for enhancing the reasoning capabilities and overall accuracy of LLMs. We categorize TTC techniques into two categories based on their resource utilization patterns: 1) techniques that increase the input tokens more~\citep{llmonk, 2022arXiv220311171W, few_shot} or 2) those that increase output tokens more~\citep{deepseek, 2023arXiv230317651M, star, rest}. Each category uniquely affects the resource demands of the inference pipeline.

Increasing the number of input tokens primarily impacts the prefill stage as the compute requirement scales quadratically with the length of the input sequence. On the other hand, increasing the number of output tokens predominantly affects the decode stage. Specifically, extending the output length linearly increases both the size of the KV-cache and the number of autoregressive decoding steps, scaling memory BW requirements quadratically.

\paragraph{Energy Usage Measurement}

Early pioneering efforts, such as the study by Strubell et al.~\citep{strubell2020energy}, began to systematically investigate the energy cost of training transformer models, spurring subsequent studies that quantified energy consumption during model training~\citep{patterson2022carbon}. With the increasing deployment of LLMs in practical applications, recent studies have expanded this scope to include energy measurements specific to inference workloads~\citep{luccioni2024power, fernandez2025energy, wu2025unveiling, patel2024characterizing, stojkovic2024towards}.

On top of the existing works, we specifically examine the additional energy overhead incurred by TTC methods. Despite TTC becoming a widely accepted practice to improve the reasoning capabilities of LLMs, the precise energy costs associated with such approaches remain largely unexplored—highlighting the critical importance and novelty of our analysis.
\section{Methodology}
\label{sec:methodology}

\begin{table}[t]
\caption{List of benchmarks used in this study for Math, Code, and Common Sense tasks}
\label{tab:tasks}
\centering
\scriptsize
\begin{tabular}{ccc}
\hline
\textbf{Math}  & \textbf{Code} & \textbf{Common Sense} \\ \hline
AIME24~\citep{aime24} & HumanEval~\citep{humaneval} & HellaSwag~\citep{hellaswag} \\
AIME25~\citep{aime25} & LiveCodeBench~\citep{livecodebench} & MMLU~\citep{mmlu} \\
GPQADiamond~\citep{gpqa}   &  MBPP~\citep{mbpp}   & GSM8K~\citep{gsm8k} \\
{Math500~\citep{math500}} & {CodeForces~\citep{codeforces}} & {CommonsenseQA~\citep{talmor-etal-2019-commonsenseqa}} \\ \hline
\end{tabular}
\end{table}

In this study, we systematically measure the additional energy consumption associated with TTC methods in the context of LLM inference. To quantify this energy overhead, we base our work on Evalchemy~\citep{evalchemy} and utilize SGLang~\citep{sglang} for model execution on NVIDIA A100 GPUs~\citep{a100}, each with a maximum power rating of 500W. We run each benchmark from start to end assuming that all questions are present to be processed to eliminate any scheduling overhead or queuing delay. Power consumption during model inference is monitored using the NVIDIA Management Library (NVML)~\citep{nvml}. We integrate power usage over the inference period to compute total energy consumption for each model evaluation.

\paragraph{Categorization of TTC Methods}
We group TTC approaches into two distinct categories: (1) methods that increase the number of input tokens and (2) methods that increase the number of output tokens. For the first category, we select parallel sampling~\citep{llmonk, getting_50, 2023arXiv230306135I, 2022arXiv220311171W} with majority vote (MV) to finalize the answer. 
For the second category, we adopt the reasoning token (RT) approach~\citep{deepseek, star, rest}. 
These models have demonstrated that they can reason through their answers by utilizing reasoning tokens.
Unless stated otherwise, we evaluate the energy usage and accuracy of MV and RT against baseline (Base) approach which does not use any TTC methodology.

\paragraph{Model Selection} We base our study on models supported by the DeepSeek-R1 framework~\citep{deepseek}, selecting four representative models from the Qwen2.5 family~\citep{qwen2.5}: 1.5B, 7B, 14B, and 32B variants. These models have been distilled from DeepSeek-R1 and are particularly suitable for demonstrating the effectiveness of reasoning tokens in refining model outputs. To facilitate a rigorous comparison, we also employ these same Qwen2.5 models as baseline references (without TTC) and in parallel sampling scenarios. Using identical model architectures across conditions ensures a fair and controlled evaluation. 32B models are executed across two NVIDIA A100 GPUs in tensor parallel manner as they on average use over 100GB compared to 80GB memory on our A100 GPUs.

\paragraph{Evaluation Tasks}
To comprehensively capture the impact of TTC, we evaluate model performance across three representative machine learning task categories, each consisting of four distinct benchmarks. We choose mathematical reasoning and code-generation tasks to assess how effectively TTC improves a model’s logical reasoning capabilities using mathematical and programming languages following existing studies~\citep{gemini, gpt4, qwen2.5}. Additionally, common sense tasks are selected to investigate whether extra computation at inference time can meaningfully enhance the model's retention or representation of factual information. Detailed descriptions of tasks and corresponding datasets are summarized in Table~\ref{tab:tasks}. 
\section{Results}
\label{sec:res}

We present our results on Base, MV, and RT. Evaluations are run 10 times with batch size 16, maximum sequence length 32768, and data type Bfloat16 unless stated otherwise. We generate 5 samples when running MV with parallel sampling. Common-sense benchmarks take roughly 10 minutes per evaluation while math and code benchmarks take between 1 to 4 hours. 

\subsection{Overview}

\begin{figure}[t]
    \centering
    \includegraphics[width=0.9\linewidth]{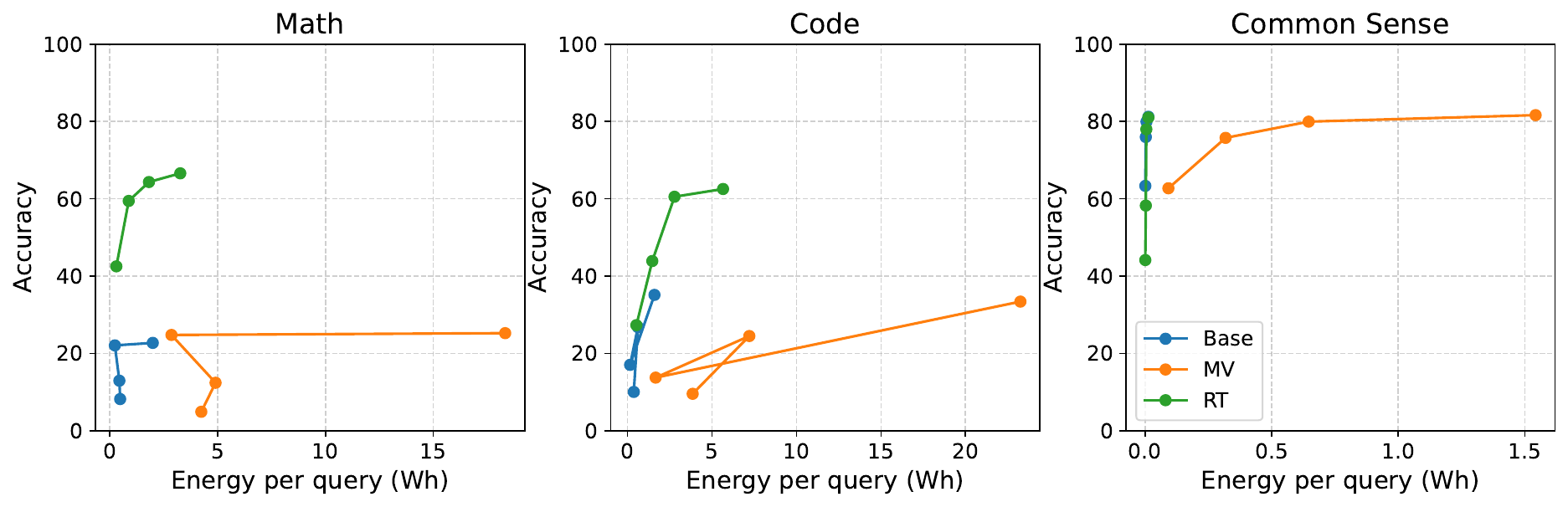}
    \caption{Accuracy versus energy per query averaged across four benchmarks in each task. Each dot in the line represent Qwen2.5 1.5B, 7B, 14B, and 32B from left to right, respectively.}
    \label{fig:main-tradeoff}
\end{figure}

\paragraph{Accuracy vs Energy per Query} The primary objective of this analysis is to explore the trade-off between accuracy improvements and associated energy costs when employing TTC. Figure~\ref{fig:main-tradeoff} presents this trade-off across mathematical, coding, and common-sense benchmarks.

Firstly, the slope of the accuracy-energy curve is steeper for math and common-sense tasks, with a clearly defined knee point situated higher and further from the origin for RT. This indicates a superior accuracy-to-energy ratio when using RT. For example, in mathematical benchmarks, increasing the model size from 1.5B to 7B parameters in the Base configuration yields only a 4.8\% accuracy improvement with negligible energy difference per query. In contrast, using RT on the 1.5B model significantly boosts accuracy by 34.3\%, representing a 7.5-fold increase in accuracy improvement over traditional scaling, at similar energy consumption levels.

In coding benchmarks, scaling model size offers better accuracy/energy gains at lower model scales than RT. Specifically, transitioning from a Base 1.5B to Base 7B model increases accuracy by 16.8\% while reducing energy consumption per query by 40.1\%. Conversely, applying RT to the 1.5B model achieves a similar accuracy improvement (17.3\%) but increases energy per query by 57.4\%. However, when targeting higher accuracy regimes (e.g., Base 32B model versus RT 7B model), RT demonstrates a superior slope of accuracy improvement per unit energy consumed (28.1\%/Wh for RT versus 20.7\%/Wh for Base). 

Common-sense benchmarks show limited or even adverse performance when employing RT. Specifically, the Base 1.5B model outperforms both the 1.5B and 7B RT configurations. This is expected because common-sense tasks predominantly evaluate factual knowledge retrieval rather than complex reasoning capabilities, limiting the effectiveness of RT.

Secondly, the use of MV combined with parallel sampling substantially increases energy consumption without proportionate accuracy improvements. This inefficiency occurs because MV relies on simplistic aggregation, where the probability of correctness, $p$, remains unchanged. Particularly, if $p < 0.5$ and we sample 5 generations, using MV with multiple candidates can potentially reduce accuracy due to the binomial distribution of generations. 
Consequently, a more efficient aggregation method is necessary to better utilize candidate outputs and justify increased energy consumption.

Finally, an intriguing anomaly emerges within the coding benchmarks: Base and MV configurations for the 14B models exhibit lower accuracy compared to their 7B counterparts. This finding reinforces the notion that merely increasing model size does not guarantee enhanced accuracy if underlying reasoning capabilities remain insufficient.

\begin{figure}[t]
    \centering
    \includegraphics[width=0.9\linewidth]{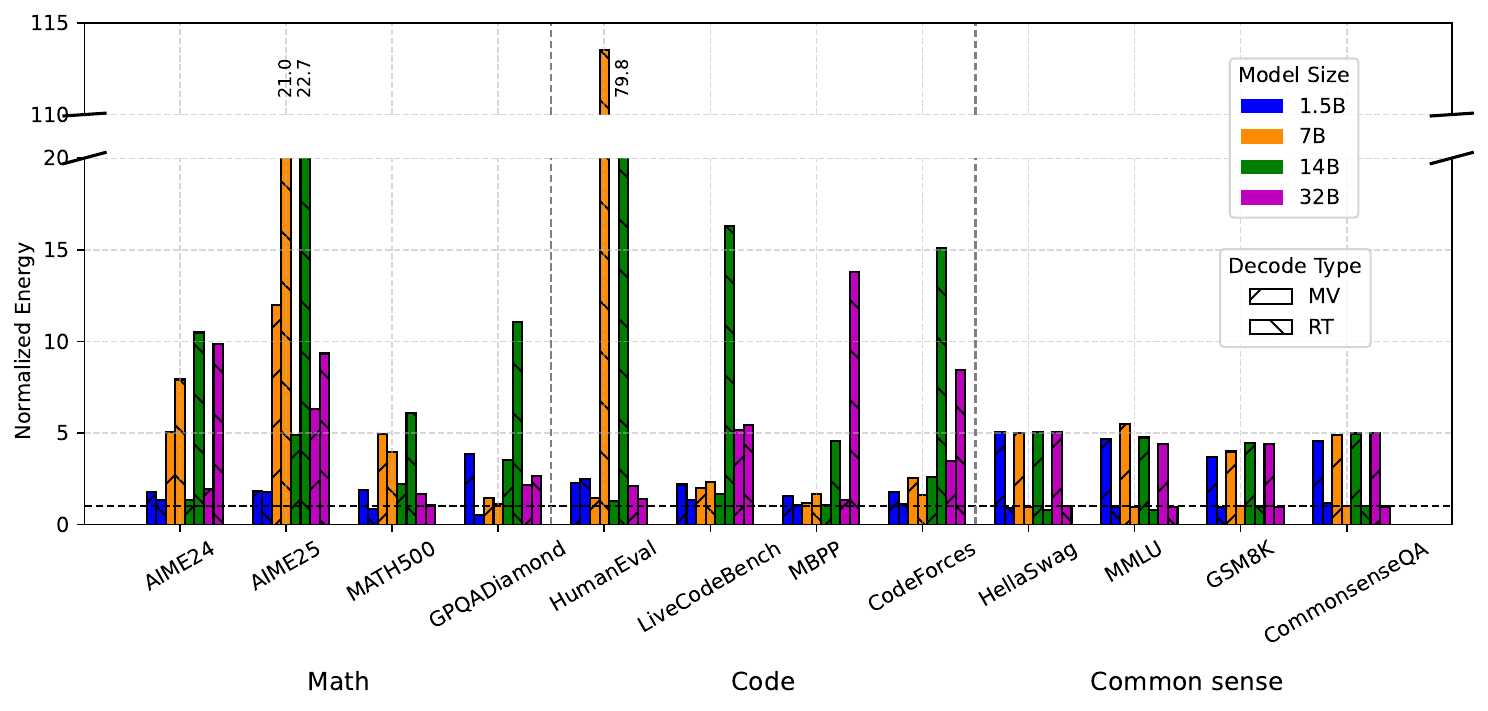}
    \caption{Energy consumption of MV and RT normalized to Base which does not use TTC. Left bars in each color represent MV and right bars represent RT. The first, second, and third sets of four benchmarks represent math, code, and common sense, respectively. The dotted horizontal line represents Base. Note that the y-axis is cut off from 20 to 110.}
    \label{fig:energy}
\end{figure}

\paragraph{Normalized Energy} We analyze the energy consumption implications of employing TTC, specifically quantifying the energy increase introduced by MV and RT relative to Base. Figure~\ref{fig:energy} illustrates these normalized energy metrics.

On average, MV consumes 2.63$\times$, 2.01$\times$, and 4.69$\times$ more energy than the Base configuration for mathematical, coding, and common-sense benchmarks, respectively. RT demonstrates a more substantial average energy increase, using 3.66$\times$, 10.4$\times$, and 1.08$\times$ more energy than Base for math, code, and common-sense tasks, respectively.

The energy cost of RT, in particular, can escalate dramatically, with a peak increase of 113.48$\times$ for the 7B model on the HumanEval coding benchmark. This extreme rise is largely driven by a substantial growth in the number of generated tokens; RT models produce, on average, 4.4$\times$ more tokens than Base, and in extreme cases up to 46.26$\times$ more tokens.

In contrast, although MV multiplies both input and output tokens by a factor of five, the short output sequences do not fully utilize GPU compute and memory BW during decoding. Consequently, this allows MV to maintain minimal latency overhead despite an increased batch size, resulting in less than a proportional 5$\times$ increase in energy consumption. However, common-sense benchmarks, which generate only a single output token correlating the answer to multiple-choice options, consistently exhibit approximately 5$\times$ energy usage relative to Base.

\begin{figure}[t]
    \centering
    \includegraphics[width=0.9\linewidth]{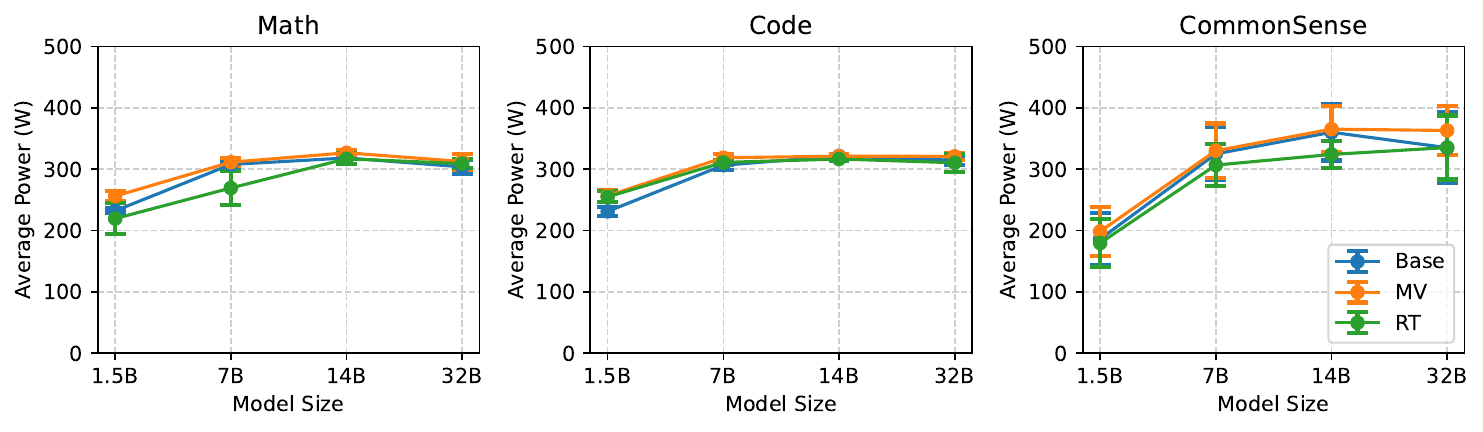}
    \caption{Power readings during runtime averaged across four benchmarks in each task.}
    \label{fig:power}
\end{figure}

\paragraph{Power Measurements} We explore power consumption implications of employing TTC. Data centers typically have strict power provisioning limits, requiring accurate assessments of power usage to prevent exceeding provisioned capacity. Figure~\ref{fig:power} provides average GPU power measurements and associated standard deviations across each benchmark.

Our analysis reveals that TTC only marginally affects power usage. Specifically, MV introduces an average increase of approximately 5\% in power consumption due to increased batch sizes from parallel sampling. Conversely, RT tends to use slightly less power than Base. This unexpected reduction occurs because RT inference is predominantly decode- and especially attention layer-bound, characterized by extremely long output sequences. The resulting high memory requirements for the KV-cache create a bottleneck that saturates GPU memory BW but significantly under-utilizes the GPU compute units.

For instance, RT benchmarks produce an average of 7845 tokens per query, translating to an average KV-cache size of about 258MB, even for relatively small (7B) models. Given the memory BW saturation threshold (approximately 100MB data array for A100 GPUs~\citep{9307857}), the inference process becomes memory-bound, limiting the effective utilization of GPU computational resources. Consequently, the power usage stabilizes at a level consistent with these memory constraints.

\subsection{In-depth Analysis}
We conclude that parallel sampling with MV shows lower accuracy/energy than RT and that TTC has limited benefit on common-sense benchmarks. To this end, we provide a thorough analysis on RT using one reasoning benchmark each from math (MATH500) and coding (LiveCodeBench) task.

\begin{figure}[t]
    \centering
    \includegraphics[width=0.9\linewidth]{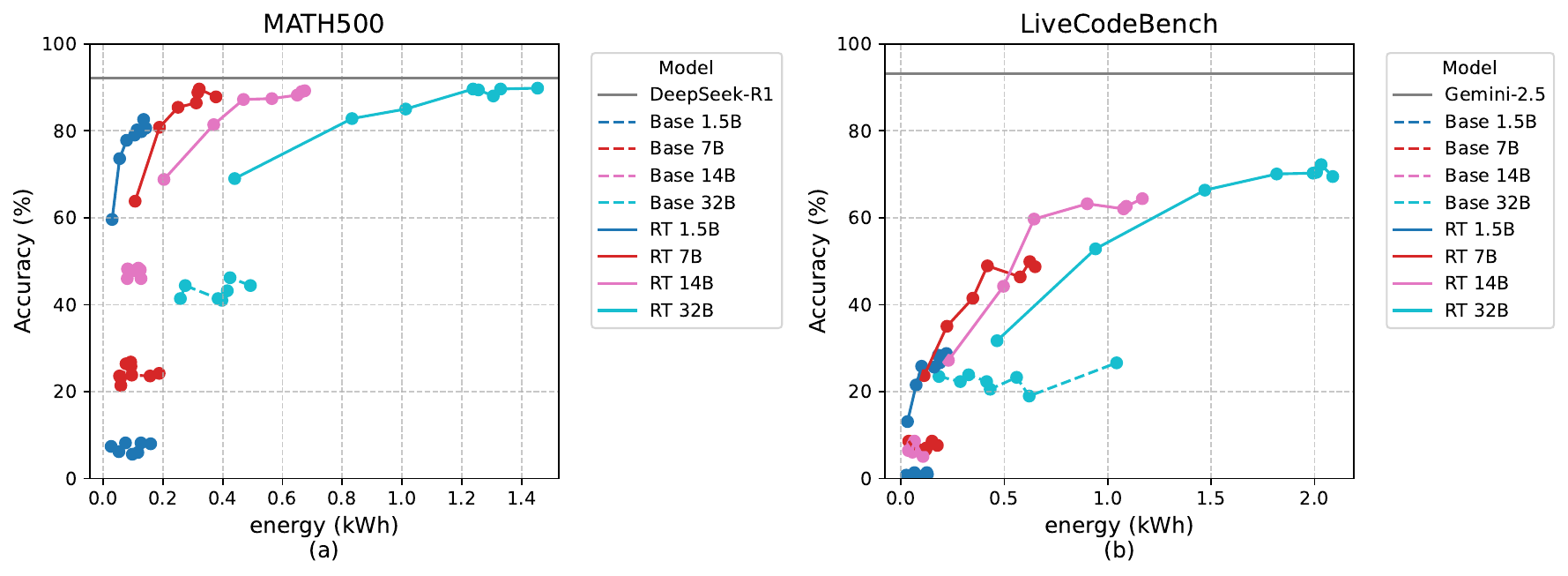}
    \caption{Energy vs Accuracy per length. Dotted and solid lines represent Base and RT, respectively and each color represent different model size. Each of the ten dots on a line represent output sequence length limit starting from one-tenth of the maximum sequence length of a model to the maximum sequence length. Grey lines on the top represent the best models at the time of writing.}
    \label{fig:length-sweep}
\end{figure}

\paragraph{Length Sweep} We investigate how the maximum output sequence length affects RT performance and associated computational costs. We incrementally limit the output sequence length by steps of 3277 tokens which is one-tenth of the original maximum sequence length.

Figure~\ref{fig:length-sweep} demonstrates the relationship between accuracy and energy consumption for two representative benchmarks. Notably, RT consistently forms a distinct and superior Pareto frontier compared to Base models. This distinction arises because Base models reach an accuracy ceiling even when allowed extended sequence lengths, with improvements requiring substantially larger model sizes. Base accuracy saturates at 48.3\% for MATH500 and  23.5\% for LiveCodeBench, whereas RT achieves significantly higher accuracy ceilings of 89.8\% and 72.1\%, respectively. Moreover, the RT Pareto frontier highlights that slight accuracy trade-offs can yield considerable energy savings, valuable in resource-constrained environments. For example, transitioning from RT 32B to RT 14B to solve LiveCodeBench reduces energy consumption by 1.74× while sacrificing only 7.8\% accuracy.

Additionally, RT models often terminate computations before reaching the maximum sequence length, which can be seen from the clustered points at the end of all RT lines. This indicates their ability to avoid excessive processing when uncertain, thus efficiently managing resources.

Figure 3(a) reveals the competitive potential of smaller RT models. For instance, the 1.5B RT model achieves 83.2\% accuracy with just 135.22Wh, outperforming the Base 32B model at only 45.7\% accuracy and consuming 424.54Wh. Remarkably, the RT 7B model reaches an accuracy of 89.6\%, approaching the performance of much larger models such as DeepSeek-R1 with 617B parameters.

\begin{figure}[t]
    \centering
    \includegraphics[width=0.9\linewidth]{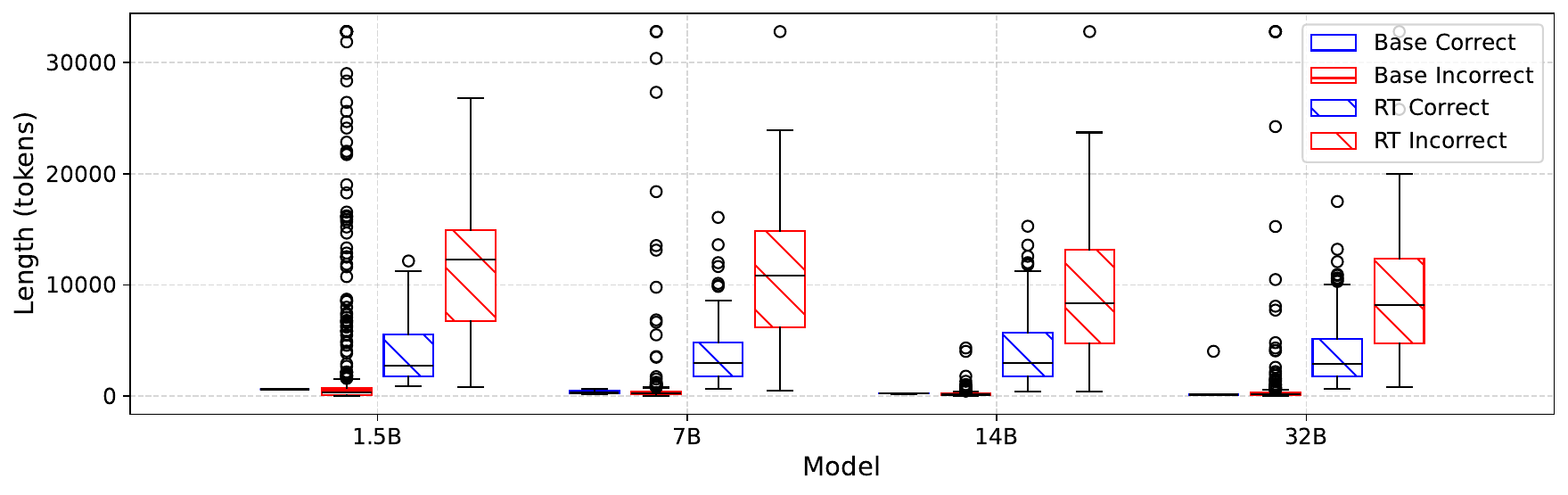}
    \caption{Output token count distribution of correct and incorrect queries.}
    \label{fig:length}
\end{figure}

\vspace{-0.3cm}
\paragraph{Tokens per Correct and Incorrect Queries} Figure~\ref{fig:length} illustrates the number of output tokens generated for correct and incorrect answers in both Base and RT models highlighting how sequence length relates to model reasoning.

Base models exhibit clustered sequence lengths for both correct and incorrect answers, with numerous outliers among incorrect responses. These models' limited reasoning capabilities result in limited thinking process and outlier sequences typically represent unnecessary token generation rather than deeper reasoning. Increasing the model size reduces the number of outliers but maintains clustering near short sequence lengths, underscoring their restricted reasoning abilities.

Conversely, RT models utilize fewer tokens to reach correct answers compared to incorrect responses. For correct queries, token length distributions are consistent across model sizes, suggesting that RT effectively guides reasoning up to a certain point. However, when uncertain, RT models produce longer incorrect responses due to loops or divergent reasoning. Larger models significantly reduce the token lengths of incorrect answers, indicating a stronger capability to recognize ineffective reasoning paths and terminate more efficiently.

\begin{figure}[t]
    \centering
    \includegraphics[width=0.9\linewidth]{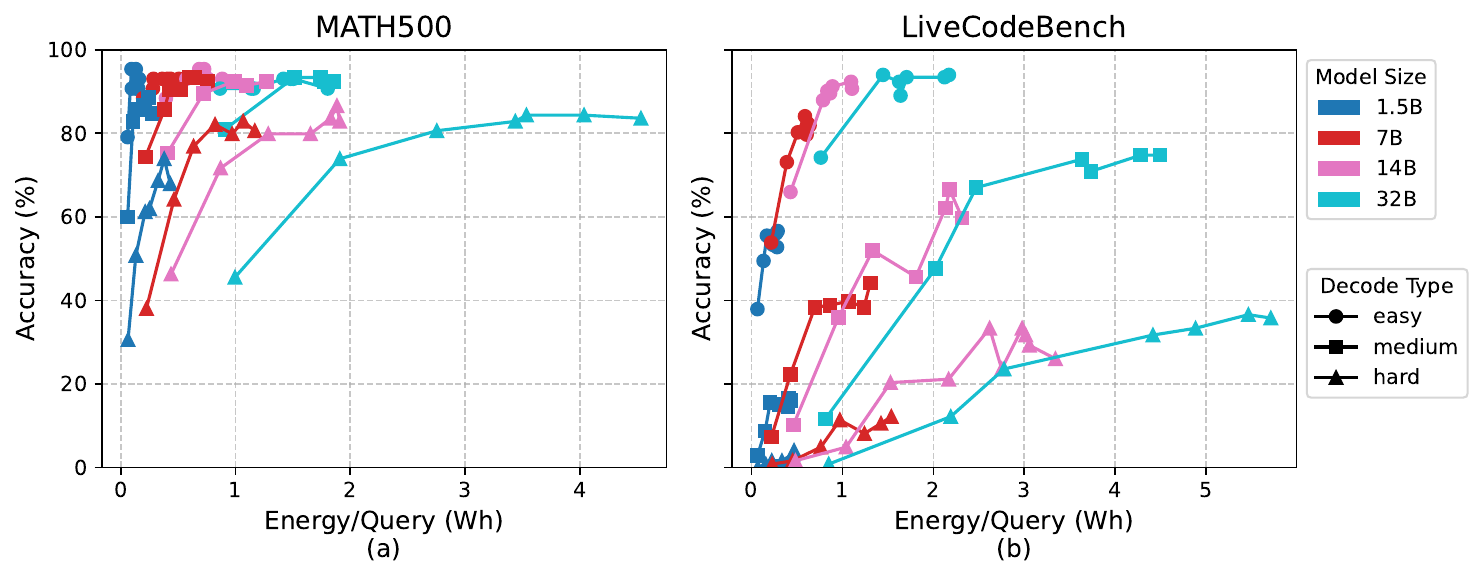}
    \caption{Accuracy vs energy per difficulty level. Each dot on a line represents different output sequence length limit identical to the previous Length Sweep analysis.}
    \label{fig:difficulty}
\end{figure}

\vspace{-0.3cm}
\paragraph{Query Difficulty} We analyze the impact of task difficulty on energy consumption and accuracy using RT. Figure~\ref{fig:difficulty} depicts the accuracy versus energy-per-query tradeoff across easy, medium, and hard difficulty levels for selected benchmarks.

Firstly, task difficulty directly correlates with increased token generation and higher energy consumption, with each difficulty level forming distinct Pareto frontiers. This indicates alignment between model-perceived and human-perceived task difficulty, where additional computational effort reflects increased complexity. This observation expands on previous findings that models' and humans' perceptions of difficulty align, especially for reasoning-based tasks.

Secondly, smaller models with RT can match the performance of larger models for easier tasks, resulting in significant energy savings. Specifically, using 14B and 1.5B models instead of 32B models saves 11.55$\times$ and 1.48$\times$ energy for easy tasks in MATH500 and LiveCodeBench, respectively.

Further analysis reveals that using RT provides better accuracy-per-energy improvements than simply increasing model size, even when all models employ RT. For example, on the hard queries of MATH500, the 14B model increases accuracy from 46.2\% to 71.6\% at an additional 0.4Wh, whereas scaling to a 32B model has marginal impact on accuracy while using 0.6Wh more energy. Similarly, the 14B model for medium-difficulty tasks in LiveCodeBench improves accuracy from 10.0\% to 36.2\% with an additional 0.96Wh of energy, demonstrating an accuracy/energy benefit of 39.5\%/Wh compared to only 3.33\%/Wh when scaling up to the 32B model. Thus, RT consistently offers superior or equivalent accuracy-per-energy tradeoffs compared to increasing model sizes.

\paragraph{Impact of Batch Size on Throughput} We examines the influence of batch size on throughput (queries-per-second, QPS) of RT compared to Base. Increasing batch size is a common strategy to enhance throughput, particularly important given the increased latency and reduced throughput associated with the longer output sequences generated by RT.

\begin{figure}[t]
    \centering
    \includegraphics[width=0.9\linewidth]{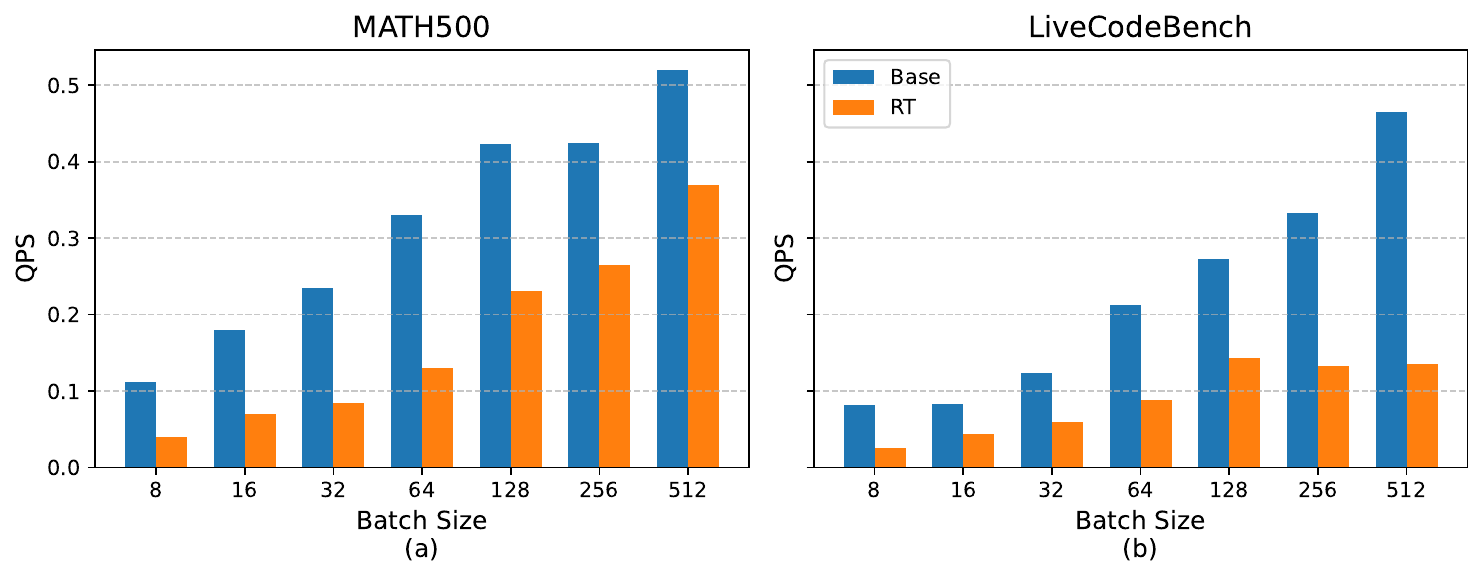}
    \caption{Throughput vs Batch size.}
    \label{fig:batch}
\end{figure}

Figure~\ref{fig:batch} depicts the QPS across various batch sizes for the two benchmarks using the highest-performing 32B models. Results indicate that larger batch sizes consistently improve throughput for both configurations. For Base, output sequences are relatively short, resulting in smaller KV-cache footprints that under-utilize GPU memory BW. Although RT applied to MATH500 generates more tokens than Base, it still generates short enough sequences to under-utilize GPU memory BW. As a result, even at high batch sizes, the KV-cache remains manageable, and throughput scales proportionally with batch size increases.

However, RT applied to the LiveCodeBench exhibit a notable saturation in throughput at a batch size of approximately 128. On average, RT generates around 6588 output tokens for LiveCodeBench, shifting inference into a predominantly decode-bound stage. During the decode stage, loading the large KV-cache that scales linearly with batch size from GPU memory becomes the critical performance bottleneck, eventually saturating available GPU memory BW. 

\paragraph{Impact of LLM Serving System Optimizations}

The most relevant optimization in our study is prefix caching, which stores the KV-cache of a query and reuses it whenever the same prefix appears in later queries. As shown in Table~\ref{tab:prefix_cache}, prefix caching offers little energy savings for RT, since RT does not use additional tokens during the prefill stage. MV shows a more nuanced picture as prefix caching does reduce energy consumption, but less than expected. In theory, the normalized energy consumption for MV should approach 1 if queries are dominated by the prefill stage. However, we find that MV generates at least one sample with an output sequence an order of magnitude longer than the other samples. This costly event shifts the bottleneck from prefill to decode, making the decode stage the primary driver of energy usage in the benchmarks we evaluate.

System optimizations targeting the decode stage such as speculative decoding~\citep{10.5555/3618408.3619203, 2024arXiv240115077L, 2023arXiv230201318C, 2023arXiv230207863K} benefits RT and MV with prefix caching more than Base. Speculative decoding reduces the cost of the decode stage for TTC and Base by the same factor~\citep{10.5555/3618408.3619203} by using a smaller model to generate multiple tokens and calling an original model only to verify the tokens. The overall benefit for MV and RT is larger than that for Base as the decode stage is more dominant than the prefill stage for the two TTC approaches. 
Another major proposal disaggregated serving~\citep{2024arXiv240109670Z}, separates prefill and decode across different clusters. This approach allows systems to reclaim idle resources, but incurs overhead from transferring the KV-cache between clusters. Importantly, just like speculative decoding, disaggregated serving applies its costs and savings uniformly to TTC and Base, maintaining the trend that we report in this study.

\begin{table}[t]
\caption{Energy consumption of MV and RT with and without the use of prefix caching normalized to Base. Normalized energy consumption for RT stays similar as no tokens were added during the prefill stage. Prefix caching saves energy consumption for MV but smaller than anticipated.}
\centering
\begin{tabular}{ccccc}
\hline
\multicolumn{1}{l}{} & \multicolumn{2}{c}{\textbf{MV}}         & \multicolumn{2}{c}{\textbf{RT}}         \\ \hline
\textbf{Tasks}                & \textbf{Prefix Cache} & \textbf{No Prefix Cache} & \textbf{Prefix Cache} & \textbf{No Prefix Cache} \\ \hline
Math                 & 1.91         & 2.28            & 2.11         & 2.37            \\
Code                 & 2.45         & 3.52            & 4.85         & 4.94            \\
Common Sense         & 4.96         & 4.75            & 1.01         & 1.01            \\ \hline
\end{tabular}
\label{tab:prefix_cache}
\end{table}
\section{Conclusion and Discussion}

In this work, we systematically evaluate the energy consumption of large language model (LLM) inference specifically through Test-Time Computation (TTC). We investigated two prevalent TTC strategies: (1) generating multiple candidate answers, and (2) self-refinement through iterative reasoning. Our analysis reveals that employing TTC strategies generally achieves better accuracy-energy trade-offs than solely increasing model size. 
Additionally, output sequence length emerges as a reliable indicator of model comprehension and complex questions require greater computational resources during inference. 
Combining larger models with TTC strategies yields the highest accuracy but comes with substantial energy costs. Thus, a prudent approach is essential, balancing accuracy gains against energy expenditures. Our findings suggest promising directions for leveraging TTC in a more energy-efficient manner.

\paragraph{Difficulty-aware Model Selection} Our results (Figure~\ref{fig:length-sweep}) demonstrate that smaller models with TTC can compete effectively against larger models, with distinct Pareto frontiers emerging for each task difficulty level. This insight supports the implementation of difficulty-aware evaluators~\citep{gpt5, 2024arXiv240407720S, park2024large, dutulescu2024hard, xu2024adaption} that dynamically route queries to the most energy-efficient model achieving optimal accuracy. Optimal energy consumption $E_o$ of an oracle evaluator and realistic energy consumption $E_r$ of an imperfect evaluator in practical settings employing a high-performing model as a fallback can be calculated as

\begin{subequations}
\begin{minipage}{0.45\textwidth}
\begin{equation}
    E_o = \sum_{i=0}^{D} n_i e_{m_i}
\end{equation}
\end{minipage}
\hfill
\begin{minipage}{0.45\textwidth}
\begin{equation}
    E_r = \sum_{i=0}^{D} p_i N e_{m_i} + q N e_{m_D} 
\end{equation}
\end{minipage}
\end{subequations}

where $p_i$ denotes the probability that the evaluator correctly predicts difficulty level $i$, $q$ represents the probability of incorrect predictions, and $e_{m_D}$ is the energy consumed by the largest, most accurate model. Compared against the baseline approach ($E_{max} = N e_{m_D}$), our analysis shows significant potential energy savings. Systems with an oracle evaluator consume 1.75 kWh compared to baseline consumptions of 2.03 kWh and 2.20 kWh, for LiveCodeBench and MATH500, respectively.

\paragraph{Length-wise Early Exit} Figure~\ref{fig:length} indicates that correct responses typically require fewer output tokens than incorrect responses, suggesting another promising avenue for efficiency improvements. Existing early-exit methods focus on exiting at certain model layers~\citep{del2023skipdecode, chen2023ee, teerapittayanon2016branchynet, zhou2020bert}. Extending these techniques to include token-length threshold will enable models to cease generation when answers become evidently flawed or excessively verbose. Such a mechanism effectively reduces unnecessary energy consumption. Specifically, Figure~\ref{fig:length} illustrates that the correct answers use less than 12000 tokens excluding outliers. We can apply this observation on Figure~\ref{fig:length-sweep}, cutting the generation at the fourth point (13108 tokens) on the RT lines. This gives us 14\% accuracy saving without loss of accuracy.


\bibliographystyle{plainnat}
\bibliography{arxiv}

@inproceedings{math500,
 author = {Hendrycks, Dan and Burns, Collin and Kadavath, Saurav and Arora, Akul and Basart, Steven and Tang, Eric and Song, Dawn and Steinhardt, Jacob},
 booktitle = {Proceedings of the Neural Information Processing Systems Track on Datasets and Benchmarks},
 editor = {J. Vanschoren and S. Yeung},
 pages = {},
 title = {Measuring Mathematical Problem Solving With the MATH Dataset},
 url = {https://datasets-benchmarks-proceedings.neurips.cc/paper_files/paper/2021/file/be83ab3ecd0db773eb2dc1b0a17836a1-Paper-round2.pdf},
 volume = {1},
 year = {2021}
}

@inproceedings{gpqa,
  title={Gpqa: A graduate-level google-proof q\&a benchmark},
  author={Rein, David and Hou, Betty Li and Stickland, Asa Cooper and Petty, Jackson and Pang, Richard Yuanzhe and Dirani, Julien and Michael, Julian and Bowman, Samuel R},
  booktitle={First Conference on Language Modeling},
  year={2024}
}

@ARTICLE{humaneval,
       author = {{Chen}, Mark and {Tworek}, Jerry and {Jun}, Heewoo and {Yuan}, Qiming and {Ponde de Oliveira Pinto}, Henrique and {Kaplan}, Jared and {Edwards}, Harri and {Burda}, Yuri and {Joseph}, Nicholas and {Brockman}, Greg and {Ray}, Alex and {Puri}, Raul and {Krueger}, Gretchen and {Petrov}, Michael and {Khlaaf}, Heidy and {Sastry}, Girish and {Mishkin}, Pamela and {Chan}, Brooke and {Gray}, Scott and {Ryder}, Nick and {Pavlov}, Mikhail and {Power}, Alethea and {Kaiser}, Lukasz and {Bavarian}, Mohammad and {Winter}, Clemens and {Tillet}, Philippe and {Petroski Such}, Felipe and {Cummings}, Dave and {Plappert}, Matthias and {Chantzis}, Fotios and {Barnes}, Elizabeth and {Herbert-Voss}, Ariel and {Hebgen Guss}, William and {Nichol}, Alex and {Paino}, Alex and {Tezak}, Nikolas and {Tang}, Jie and {Babuschkin}, Igor and {Balaji}, Suchir and {Jain}, Shantanu and {Saunders}, William and {Hesse}, Christopher and {Carr}, Andrew N. and {Leike}, Jan and {Achiam}, Josh and {Misra}, Vedant and {Morikawa}, Evan and {Radford}, Alec and {Knight}, Matthew and {Brundage}, Miles and {Murati}, Mira and {Mayer}, Katie and {Welinder}, Peter and {McGrew}, Bob and {Amodei}, Dario and {McCandlish}, Sam and {Sutskever}, Ilya and {Zaremba}, Wojciech},
        title = "{Evaluating Large Language Models Trained on Code}",
      journal = {arXiv e-prints},
         year = 2021,
        month = jul,
          eid = {arXiv:2107.03374},
        pages = {arXiv:2107.03374},
          doi = {10.48550/arXiv.2107.03374},
archivePrefix = {arXiv},
       eprint = {2107.03374},
 primaryClass = {cs.LG},
       adsurl = {https://ui.adsabs.harvard.edu/abs/2021arXiv210703374C}
}

@ARTICLE{livecodebench,
       author = {{Jain}, Naman and {Han}, King and {Gu}, Alex and {Li}, Wen-Ding and {Yan}, Fanjia and {Zhang}, Tianjun and {Wang}, Sida and {Solar-Lezama}, Armando and {Sen}, Koushik and {Stoica}, Ion},
        title = "{LiveCodeBench: Holistic and Contamination Free Evaluation of Large Language Models for Code}",
      journal = {arXiv e-prints},
         year = 2024,
        month = mar,
          eid = {arXiv:2403.07974},
        pages = {arXiv:2403.07974},
          doi = {10.48550/arXiv.2403.07974},
archivePrefix = {arXiv},
       eprint = {2403.07974},
 primaryClass = {cs.SE},
       adsurl = {https://ui.adsabs.harvard.edu/abs/2024arXiv240307974J}
}

@ARTICLE{mbpp,
       author = {{Austin}, Jacob and {Odena}, Augustus and {Nye}, Maxwell and {Bosma}, Maarten and {Michalewski}, Henryk and {Dohan}, David and {Jiang}, Ellen and {Cai}, Carrie and {Terry}, Michael and {Le}, Quoc and {Sutton}, Charles},
        title = "{Program Synthesis with Large Language Models}",
      journal = {arXiv e-prints},
         year = 2021,
        month = aug,
          eid = {arXiv:2108.07732},
        pages = {arXiv:2108.07732},
          doi = {10.48550/arXiv.2108.07732},
archivePrefix = {arXiv},
       eprint = {2108.07732},
 primaryClass = {cs.PL},
       adsurl = {https://ui.adsabs.harvard.edu/abs/2021arXiv210807732A}
}

@ARTICLE{hellaswag,
       author = {{Zellers}, Rowan and {Holtzman}, Ari and {Bisk}, Yonatan and {Farhadi}, Ali and {Choi}, Yejin},
        title = "{HellaSwag: Can a Machine Really Finish Your Sentence?}",
      journal = {arXiv e-prints},
         year = 2019,
        month = may,
          eid = {arXiv:1905.07830},
        pages = {arXiv:1905.07830},
          doi = {10.48550/arXiv.1905.07830},
archivePrefix = {arXiv},
       eprint = {1905.07830},
 primaryClass = {cs.CL},
       adsurl = {https://ui.adsabs.harvard.edu/abs/2019arXiv190507830Z}
}

@ARTICLE{gsm8k,
       author = {{Cobbe}, Karl and {Kosaraju}, Vineet and {Bavarian}, Mohammad and {Chen}, Mark and {Jun}, Heewoo and {Kaiser}, Lukasz and {Plappert}, Matthias and {Tworek}, Jerry and {Hilton}, Jacob and {Nakano}, Reiichiro and {Hesse}, Christopher and {Schulman}, John},
        title = "{Training Verifiers to Solve Math Word Problems}",
      journal = {arXiv e-prints},
         year = 2021,
        month = oct,
          eid = {arXiv:2110.14168},
        pages = {arXiv:2110.14168},
          doi = {10.48550/arXiv.2110.14168},
archivePrefix = {arXiv},
       eprint = {2110.14168},
 primaryClass = {cs.LG},
       adsurl = {https://ui.adsabs.harvard.edu/abs/2021arXiv211014168C}
}

@article{mmlu,
  title={Measuring Massive Multitask Language Understanding},
  author={Dan Hendrycks and Collin Burns and Steven Basart and Andy Zou and Mantas Mazeika and Dawn Song and Jacob Steinhardt},
  journal={Proceedings of the International Conference on Learning Representations (ICLR)},
  year={2021}
}

@misc{a100,
  author = {NVIDIA},
  title = {{NVIDIA A100 GPU}},
  howpublished = {\url{https://www.nvidia.com/content/dam/en-zz/Solutions/Data-Center/a100/pdf/nvidia-a100-datasheet-nvidia-us-2188504-web.pdf}},
  year = {2020}
}

@misc{nvml,
  author = {NVIDIA},
  title = {{NVIDIA NVML Documentation}},
  howpublished = {https://developer.nvidia.com/management-library-nvml},
  year = {2024}
}

@ARTICLE{deepseek,
       author = {{DeepSeek-AI} and {Guo}, Daya and {Yang}, Dejian and {Zhang}, Haowei and {Song}, Junxiao and {Zhang}, Ruoyu and {Xu}, Runxin and {Zhu}, Qihao and {Ma}, Shirong and {Wang}, Peiyi and {Bi}, Xiao and {Zhang}, Xiaokang and {Yu}, Xingkai and {Wu}, Yu and {Wu}, Z.~F. and {Gou}, Zhibin and {Shao}, Zhihong and {Li}, Zhuoshu and {Gao}, Ziyi and {Liu}, Aixin and {Xue}, Bing and {Wang}, Bingxuan and {Wu}, Bochao and {Feng}, Bei and {Lu}, Chengda and {Zhao}, Chenggang and {Deng}, Chengqi and {Zhang}, Chenyu and {Ruan}, Chong and {Dai}, Damai and {Chen}, Deli and {Ji}, Dongjie and {Li}, Erhang and {Lin}, Fangyun and {Dai}, Fucong and {Luo}, Fuli and {Hao}, Guangbo and {Chen}, Guanting and {Li}, Guowei and {Zhang}, H. and {Bao}, Han and {Xu}, Hanwei and {Wang}, Haocheng and {Ding}, Honghui and {Xin}, Huajian and {Gao}, Huazuo and {Qu}, Hui and {Li}, Hui and {Guo}, Jianzhong and {Li}, Jiashi and {Wang}, Jiawei and {Chen}, Jingchang and {Yuan}, Jingyang and {Qiu}, Junjie and {Li}, Junlong and {Cai}, J.~L. and {Ni}, Jiaqi and {Liang}, Jian and {Chen}, Jin and {Dong}, Kai and {Hu}, Kai and {Gao}, Kaige and {Guan}, Kang and {Huang}, Kexin and {Yu}, Kuai and {Wang}, Lean and {Zhang}, Lecong and {Zhao}, Liang and {Wang}, Litong and {Zhang}, Liyue and {Xu}, Lei and {Xia}, Leyi and {Zhang}, Mingchuan and {Zhang}, Minghua and {Tang}, Minghui and {Li}, Meng and {Wang}, Miaojun and {Li}, Mingming and {Tian}, Ning and {Huang}, Panpan and {Zhang}, Peng and {Wang}, Qiancheng and {Chen}, Qinyu and {Du}, Qiushi and {Ge}, Ruiqi and {Zhang}, Ruisong and {Pan}, Ruizhe and {Wang}, Runji and {Chen}, R.~J. and {Jin}, R.~L. and {Chen}, Ruyi and {Lu}, Shanghao and {Zhou}, Shangyan and {Chen}, Shanhuang and {Ye}, Shengfeng and {Wang}, Shiyu and {Yu}, Shuiping and {Zhou}, Shunfeng and {Pan}, Shuting and {Li}, S.~S. and {Zhou}, Shuang and {Wu}, Shaoqing and {Ye}, Shengfeng and {Yun}, Tao and {Pei}, Tian and {Sun}, Tianyu and {Wang}, T. and {Zeng}, Wangding and {Zhao}, Wanjia and {Liu}, Wen and {Liang}, Wenfeng and {Gao}, Wenjun and {Yu}, Wenqin and {Zhang}, Wentao and {Xiao}, W.~L. and {An}, Wei and {Liu}, Xiaodong and {Wang}, Xiaohan and {Chen}, Xiaokang and {Nie}, Xiaotao and {Cheng}, Xin and {Liu}, Xin and {Xie}, Xin and {Liu}, Xingchao and {Yang}, Xinyu and {Li}, Xinyuan and {Su}, Xuecheng and {Lin}, Xuheng and {Li}, X.~Q. and {Jin}, Xiangyue and {Shen}, Xiaojin and {Chen}, Xiaosha and {Sun}, Xiaowen and {Wang}, Xiaoxiang and {Song}, Xinnan and {Zhou}, Xinyi and {Wang}, Xianzu and {Shan}, Xinxia and {Li}, Y.~K. and {Wang}, Y.~Q. and {Wei}, Y.~X. and {Zhang}, Yang and {Xu}, Yanhong and {Li}, Yao and {Zhao}, Yao and {Sun}, Yaofeng and {Wang}, Yaohui and {Yu}, Yi and {Zhang}, Yichao and {Shi}, Yifan and {Xiong}, Yiliang and {He}, Ying and {Piao}, Yishi and {Wang}, Yisong and {Tan}, Yixuan and {Ma}, Yiyang and {Liu}, Yiyuan and {Guo}, Yongqiang and {Ou}, Yuan and {Wang}, Yuduan and {Gong}, Yue and {Zou}, Yuheng and {He}, Yujia and {Xiong}, Yunfan and {Luo}, Yuxiang and {You}, Yuxiang and {Liu}, Yuxuan and {Zhou}, Yuyang and {Zhu}, Y.~X. and {Xu}, Yanhong and {Huang}, Yanping and {Li}, Yaohui and {Zheng}, Yi and {Zhu}, Yuchen and {Ma}, Yunxian and {Tang}, Ying and {Zha}, Yukun and {Yan}, Yuting and {Ren}, Z.~Z. and {Ren}, Zehui and {Sha}, Zhangli and {Fu}, Zhe and {Xu}, Zhean and {Xie}, Zhenda and {Zhang}, Zhengyan and {Hao}, Zhewen and {Ma}, Zhicheng and {Yan}, Zhigang and {Wu}, Zhiyu and {Gu}, Zihui and {Zhu}, Zijia and {Liu}, Zijun and {Li}, Zilin and {Xie}, Ziwei and {Song}, Ziyang and {Pan}, Zizheng and {Huang}, Zhen and {Xu}, Zhipeng and {Zhang}, Zhongyu and {Zhang}, Zhen},
        title = "{DeepSeek-R1: Incentivizing Reasoning Capability in LLMs via Reinforcement Learning}",
      journal = {arXiv e-prints},
         year = 2025,
        month = jan,
          eid = {arXiv:2501.12948},
        pages = {arXiv:2501.12948},
          doi = {10.48550/arXiv.2501.12948},
archivePrefix = {arXiv},
       eprint = {2501.12948},
 primaryClass = {cs.CL},
       adsurl = {https://ui.adsabs.harvard.edu/abs/2025arXiv250112948D}
}

@ARTICLE{llmonk,
       author = {{Brown}, Bradley and {Juravsky}, Jordan and {Ehrlich}, Ryan and {Clark}, Ronald and {Le}, Quoc V. and {R{\'e}}, Christopher and {Mirhoseini}, Azalia},
        title = "{Large Language Monkeys: Scaling Inference Compute with Repeated Sampling}",
      journal = {arXiv e-prints},
         year = 2024,
        month = jul,
          eid = {arXiv:2407.21787},
        pages = {arXiv:2407.21787},
          doi = {10.48550/arXiv.2407.21787},
archivePrefix = {arXiv},
       eprint = {2407.21787},
 primaryClass = {cs.LG},
       adsurl = {https://ui.adsabs.harvard.edu/abs/2024arXiv240721787B}
}

@inproceedings{sglang,
 author = {Zheng, Lianmin and Yin, Liangsheng and Xie, Zhiqiang and Sun, Chuyue and Huang, Jeff and Yu, Cody Hao and Cao, Shiyi and Kozyrakis, Christos and Stoica, Ion and Gonzalez, Joseph E. and Barrett, Clark and Sheng, Ying},
 booktitle = {Advances in Neural Information Processing Systems},
 editor = {A. Globerson and L. Mackey and D. Belgrave and A. Fan and U. Paquet and J. Tomczak and C. Zhang},
 pages = {62557--62583},
 publisher = {Curran Associates, Inc.},
 title = {SGLang: Efficient Execution of Structured Language Model Programs},
 url = {https://proceedings.neurips.cc/paper_files/paper/2024/file/724be4472168f31ba1c9ac630f15dec8-Paper-Conference.pdf},
 volume = {37},
 year = {2024}
}

@misc{aime24,
  title={{AIME24}},
  author={{MAA}},
  howpublished={https://maa.org/math
-competitions/american-invitational-mathematics-examination-aime},
  journal={American invitational mathematics examination - aime. In American Invitational Mathematics Examination - AIME 2024},
  year={February 2024}
}

@misc{aime25,
  title={{AIME25}},
  author={{MAA}},
  howpublished={https://maa.org/math
-competitions/american-invitational-mathematics-examination-aime},
  journal={American invitational mathematics examination - aime. In American Invitational Mathematics Examination - AIME 2025},
  year={February 2024}
}

@ARTICLE{codeforces,
       author = {{Quan}, Shanghaoran and {Yang}, Jiaxi and {Yu}, Bowen and {Zheng}, Bo and {Liu}, Dayiheng and {Yang}, An and {Ren}, Xuancheng and {Gao}, Bofei and {Miao}, Yibo and {Feng}, Yunlong and {Wang}, Zekun and {Yang}, Jian and {Cui}, Zeyu and {Fan}, Yang and {Zhang}, Yichang and {Hui}, Binyuan and {Lin}, Junyang},
        title = "{CodeElo: Benchmarking Competition-level Code Generation of LLMs with Human-comparable Elo Ratings}",
      journal = {arXiv e-prints},
         year = 2025,
        month = jan,
          eid = {arXiv:2501.01257},
        pages = {arXiv:2501.01257},
          doi = {10.48550/arXiv.2501.01257},
archivePrefix = {arXiv},
       eprint = {2501.01257},
 primaryClass = {cs.CL},
       adsurl = {https://ui.adsabs.harvard.edu/abs/2025arXiv250101257Q}
}

@INPROCEEDINGS{9307857,
  author={Anzt, Hartwig and Tsai, Yuhsiang M. and Abdelfattah, Ahmad and Cojean, Terry and Dongarra, Jack},
  booktitle={2020 IEEE/ACM Performance Modeling, Benchmarking and Simulation of High Performance Computer Systems (PMBS)}, 
  title={Evaluating the Performance of NVIDIA’s A100 Ampere GPU for Sparse and Batched Computations}, 
  year={2020},
  volume={},
  number={},
  pages={26-38},
  doi={10.1109/PMBS51919.2020.00009}}

@ARTICLE{2024arXiv240407720S,
       author = {{S{\"a}uberli}, Andreas and {Clematide}, Simon},
        title = "{Automatic Generation and Evaluation of Reading Comprehension Test Items with Large Language Models}",
      journal = {arXiv e-prints},
         year = 2024,
        month = apr,
          eid = {arXiv:2404.07720},
        pages = {arXiv:2404.07720},
          doi = {10.48550/arXiv.2404.07720},
archivePrefix = {arXiv},
       eprint = {2404.07720},
 primaryClass = {cs.CL},
       adsurl = {https://ui.adsabs.harvard.edu/abs/2024arXiv240407720S}
}

@ARTICLE{qwen2.5,
       author = {{Qwen} and {:} and {Yang}, An and {Yang}, Baosong and {Zhang}, Beichen and {Hui}, Binyuan and {Zheng}, Bo and {Yu}, Bowen and {Li}, Chengyuan and {Liu}, Dayiheng and {Huang}, Fei and {Wei}, Haoran and {Lin}, Huan and {Yang}, Jian and {Tu}, Jianhong and {Zhang}, Jianwei and {Yang}, Jianxin and {Yang}, Jiaxi and {Zhou}, Jingren and {Lin}, Junyang and {Dang}, Kai and {Lu}, Keming and {Bao}, Keqin and {Yang}, Kexin and {Yu}, Le and {Li}, Mei and {Xue}, Mingfeng and {Zhang}, Pei and {Zhu}, Qin and {Men}, Rui and {Lin}, Runji and {Li}, Tianhao and {Tang}, Tianyi and {Xia}, Tingyu and {Ren}, Xingzhang and {Ren}, Xuancheng and {Fan}, Yang and {Su}, Yang and {Zhang}, Yichang and {Wan}, Yu and {Liu}, Yuqiong and {Cui}, Zeyu and {Zhang}, Zhenru and {Qiu}, Zihan},
        title = "{Qwen2.5 Technical Report}",
      journal = {arXiv e-prints},
         year = 2024,
        month = dec,
          eid = {arXiv:2412.15115},
        pages = {arXiv:2412.15115},
          doi = {10.48550/arXiv.2412.15115},
archivePrefix = {arXiv},
       eprint = {2412.15115},
 primaryClass = {cs.CL},
       adsurl = {https://ui.adsabs.harvard.edu/abs/2024arXiv241215115Q}
}

@misc{getting_50,
  author = {Ryan Greenblatt},
  title = {{Geting 50}},
  howpublished = {\url{https://www.lesswrong.com/posts/Rdwui3wHxCeKb7feK/getting-50-sota-on-arc-agi-with-gpt-4o}},
  year = {2024}
}

@ARTICLE{2023arXiv230306135I,
       author = {{Irvine}, Robert and {Boubert}, Douglas and {Raina}, Vyas and {Liusie}, Adian and {Zhu}, Ziyi and {Mudupalli}, Vineet and {Korshuk}, Aliaksei and {Liu}, Zongyi and {Cremer}, Fritz and {Assassi}, Valentin and {Beauchamp}, Christie-Carol and {Lu}, Xiaoding and {Rialan}, Thomas and {Beauchamp}, William},
        title = "{Rewarding Chatbots for Real-World Engagement with Millions of Users}",
      journal = {arXiv e-prints},
         year = 2023,
        month = mar,
          eid = {arXiv:2303.06135},
        pages = {arXiv:2303.06135},
          doi = {10.48550/arXiv.2303.06135},
archivePrefix = {arXiv},
       eprint = {2303.06135},
 primaryClass = {cs.CL},
       adsurl = {https://ui.adsabs.harvard.edu/abs/2023arXiv230306135I}
}

@ARTICLE{2022arXiv220311171W,
       author = {{Wang}, Xuezhi and {Wei}, Jason and {Schuurmans}, Dale and {Le}, Quoc and {Chi}, Ed and {Narang}, Sharan and {Chowdhery}, Aakanksha and {Zhou}, Denny},
        title = "{Self-Consistency Improves Chain of Thought Reasoning in Language Models}",
      journal = {arXiv e-prints},
         year = 2022,
        month = mar,
          eid = {arXiv:2203.11171},
        pages = {arXiv:2203.11171},
          doi = {10.48550/arXiv.2203.11171},
archivePrefix = {arXiv},
       eprint = {2203.11171},
 primaryClass = {cs.CL},
       adsurl = {https://ui.adsabs.harvard.edu/abs/2022arXiv220311171W}
}

@ARTICLE{2022arXiv221104325V,
       author = {{Villalobos}, Pablo and {Ho}, Anson and {Sevilla}, Jaime and {Besiroglu}, Tamay and {Heim}, Lennart and {Hobbhahn}, Marius},
        title = "{Will we run out of data? Limits of LLM scaling based on human-generated data}",
      journal = {arXiv e-prints},
         year = 2022,
        month = oct,
          eid = {arXiv:2211.04325},
        pages = {arXiv:2211.04325},
          doi = {10.48550/arXiv.2211.04325},
archivePrefix = {arXiv},
       eprint = {2211.04325},
 primaryClass = {cs.LG},
       adsurl = {https://ui.adsabs.harvard.edu/abs/2022arXiv221104325V}
}

@techreport{shehabi2024united,
  title        = {2024 United States Data Center Energy Usage Report},
  author       = {Shehabi, Arman and Smith, Sarah Josephine and Hubbard, Alex and Newkirk, Alexander and Lei, Nuoa and Siddik, Md AbuBakar and Holecek, Billie and Koomey, Jonathan G and Masanet, Eric R and Sartor, Dale A},
  institution  = {Lawrence Berkeley National Laboratory},
  type         = {Report},
  number       = {LBNL-2001637},
  year         = {2024},
  month        = dec,
  doi          = {10.71468/P1WC7Q},
  url          = {https://eta-publications.lbl.gov/sites/default/files/2024-12/lbnl-2024-united-states-data-center-energy-usage-report.pdf},
}

@ARTICLE{2023arXiv230510601Y,
       author = {{Yao}, Shunyu and {Yu}, Dian and {Zhao}, Jeffrey and {Shafran}, Izhak and {Griffiths}, Thomas L. and {Cao}, Yuan and {Narasimhan}, Karthik},
        title = "{Tree of Thoughts: Deliberate Problem Solving with Large Language Models}",
      journal = {arXiv e-prints},
         year = 2023,
        month = may,
          eid = {arXiv:2305.10601},
        pages = {arXiv:2305.10601},
          doi = {10.48550/arXiv.2305.10601},
archivePrefix = {arXiv},
       eprint = {2305.10601},
 primaryClass = {cs.CL},
       adsurl = {https://ui.adsabs.harvard.edu/abs/2023arXiv230510601Y}
}

@ARTICLE{2023arXiv230317651M,
       author = {{Madaan}, Aman and {Tandon}, Niket and {Gupta}, Prakhar and {Hallinan}, Skyler and {Gao}, Luyu and {Wiegreffe}, Sarah and {Alon}, Uri and {Dziri}, Nouha and {Prabhumoye}, Shrimai and {Yang}, Yiming and {Gupta}, Shashank and {Prasad Majumder}, Bodhisattwa and {Hermann}, Katherine and {Welleck}, Sean and {Yazdanbakhsh}, Amir and {Clark}, Peter},
        title = "{Self-Refine: Iterative Refinement with Self-Feedback}",
      journal = {arXiv e-prints},
         year = 2023,
        month = mar,
          eid = {arXiv:2303.17651},
        pages = {arXiv:2303.17651},
          doi = {10.48550/arXiv.2303.17651},
archivePrefix = {arXiv},
       eprint = {2303.17651},
 primaryClass = {cs.CL},
       adsurl = {https://ui.adsabs.harvard.edu/abs/2023arXiv230317651M}
}

@ARTICLE{2024arXiv240615673L,
       author = {{Liu}, Dancheng and {Nassereldine}, Amir and {Yang}, Ziming and {Xu}, Chenhui and {Hu}, Yuting and {Li}, Jiajie and {Kumar}, Utkarsh and {Lee}, Changjae and {Qin}, Ruiyang and {Shi}, Yiyu and {Xiong}, Jinjun},
        title = "{Large Language Models have Intrinsic Self-Correction Ability}",
      journal = {arXiv e-prints},
         year = 2024,
        month = jun,
          eid = {arXiv:2406.15673},
        pages = {arXiv:2406.15673},
          doi = {10.48550/arXiv.2406.15673},
archivePrefix = {arXiv},
       eprint = {2406.15673},
 primaryClass = {cs.CL},
       adsurl = {https://ui.adsabs.harvard.edu/abs/2024arXiv240615673L}
}

@ARTICLE{2024arXiv240601297K,
       author = {{Kamoi}, Ryo and {Zhang}, Yusen and {Zhang}, Nan and {Han}, Jiawei and {Zhang}, Rui},
        title = "{When Can LLMs Actually Correct Their Own Mistakes? A Critical Survey of Self-Correction of LLMs}",
      journal = {arXiv e-prints},
         year = 2024,
        month = jun,
          eid = {arXiv:2406.01297},
        pages = {arXiv:2406.01297},
          doi = {10.48550/arXiv.2406.01297},
archivePrefix = {arXiv},
       eprint = {2406.01297},
 primaryClass = {cs.CL},
       adsurl = {https://ui.adsabs.harvard.edu/abs/2024arXiv240601297K}
}

@article{wu2022sustainable,
  title={Sustainable ai: Environmental implications, challenges and opportunities},
  author={Wu, Carole-Jean and Raghavendra, Ramya and Gupta, Udit and Acun, Bilge and Ardalani, Newsha and Maeng, Kiwan and Chang, Gloria and Aga, Fiona and Huang, Jinshi and Bai, Charles and others},
  journal={Proceedings of Machine Learning and Systems},
  volume={4},
  pages={795--813},
  year={2022}
}

@article{patterson2022carbon,
  title={The carbon footprint of machine learning training will plateau, then shrink},
  author={Patterson, David and Gonzalez, Joseph and H{\"o}lzle, Urs and Le, Quoc and Liang, Chen and Munguia, Lluis-Miquel and Rothchild, Daniel and So, David R and Texier, Maud and Dean, Jeff},
  journal={Computer},
  volume={55},
  number={7},
  pages={18--28},
  year={2022},
  publisher={IEEE}
}

@article{barr2019amazon,
  title={Amazon ec2 update--inf1 instances with AWS inferentia chips for high performance cost-effective inferencing},
  author={Barr, Jeff},
  journal={AWS News Blog},
  year={2019}
}

@ARTICLE{llama,
       author = {{Grattafiori}, Aaron and {Dubey}, Abhimanyu and {Jauhri}, Abhinav and {Pandey}, Abhinav and {Kadian}, Abhishek and {Al-Dahle}, Ahmad and {Letman}, Aiesha and {Mathur}, Akhil and {Schelten}, Alan and {Vaughan}, Alex and {Yang}, Amy and {Fan}, Angela and {Goyal}, Anirudh and {Hartshorn}, Anthony and {Yang}, Aobo and {Mitra}, Archi and {Sravankumar}, Archie and {Korenev}, Artem and {Hinsvark}, Arthur and {Rao}, Arun and {Zhang}, Aston and {Rodriguez}, Aurelien and {Gregerson}, Austen and {Spataru}, Ava and {Roziere}, Baptiste and {Biron}, Bethany and {Tang}, Binh and {Chern}, Bobbie and {Caucheteux}, Charlotte and {Nayak}, Chaya and {Bi}, Chloe and {Marra}, Chris and {McConnell}, Chris and {Keller}, Christian and {Touret}, Christophe and {Wu}, Chunyang and {Wong}, Corinne and {Canton Ferrer}, Cristian and {Nikolaidis}, Cyrus and {Allonsius}, Damien and {Song}, Daniel and {Pintz}, Danielle and {Livshits}, Danny and {Wyatt}, Danny and {Esiobu}, David and {Choudhary}, Dhruv and {Mahajan}, Dhruv and {Garcia-Olano}, Diego and {Perino}, Diego and {Hupkes}, Dieuwke and {Lakomkin}, Egor and {AlBadawy}, Ehab and {Lobanova}, Elina and {Dinan}, Emily and {Smith}, Eric Michael and {Radenovic}, Filip and {Guzm{\'a}n}, Francisco and {Zhang}, Frank and {Synnaeve}, Gabriel and {Lee}, Gabrielle and {Anderson}, Georgia Lewis and {Thattai}, Govind and {Nail}, Graeme and {Mialon}, Gregoire and {Pang}, Guan and {Cucurell}, Guillem and {Nguyen}, Hailey and {Korevaar}, Hannah and {Xu}, Hu and {Touvron}, Hugo and {Zarov}, Iliyan and {Arrieta Ibarra}, Imanol and {Kloumann}, Isabel and {Misra}, Ishan and {Evtimov}, Ivan and {Zhang}, Jack and {Copet}, Jade and {Lee}, Jaewon and {Geffert}, Jan and {Vranes}, Jana and {Park}, Jason and {Mahadeokar}, Jay and {Shah}, Jeet and {van der Linde}, Jelmer and {Billock}, Jennifer and {Hong}, Jenny and {Lee}, Jenya and {Fu}, Jeremy and {Chi}, Jianfeng and {Huang}, Jianyu and {Liu}, Jiawen and {Wang}, Jie and {Yu}, Jiecao and {Bitton}, Joanna and {Spisak}, Joe and {Park}, Jongsoo and {Rocca}, Joseph and {Johnstun}, Joshua and {Saxe}, Joshua and {Jia}, Junteng and {Vasuden Alwala}, Kalyan and {Prasad}, Karthik and {Upasani}, Kartikeya and {Plawiak}, Kate and {Li}, Ke and {Heafield}, Kenneth and {Stone}, Kevin and {El-Arini}, Khalid and {Iyer}, Krithika and {Malik}, Kshitiz and {Chiu}, Kuenley and {Bhalla}, Kunal and {Lakhotia}, Kushal and {Rantala-Yeary}, Lauren and {van der Maaten}, Laurens and {Chen}, Lawrence and {Tan}, Liang and {Jenkins}, Liz and {Martin}, Louis and {Madaan}, Lovish and {Malo}, Lubo and {Blecher}, Lukas and {Landzaat}, Lukas and {de Oliveira}, Luke and {Muzzi}, Madeline and {Pasupuleti}, Mahesh and {Singh}, Mannat and {Paluri}, Manohar and {Kardas}, Marcin and {Tsimpoukelli}, Maria and {Oldham}, Mathew and {Rita}, Mathieu and {Pavlova}, Maya and {Kambadur}, Melanie and {Lewis}, Mike and {Si}, Min and {Singh}, Mitesh Kumar and {Hassan}, Mona and {Goyal}, Naman and {Torabi}, Narjes and {Bashlykov}, Nikolay and {Bogoychev}, Nikolay and {Chatterji}, Niladri and {Zhang}, Ning and {Duchenne}, Olivier and {{\c{C}}elebi}, Onur and {Alrassy}, Patrick and {Zhang}, Pengchuan and {Li}, Pengwei and {Vasic}, Petar and {Weng}, Peter and {Bhargava}, Prajjwal and {Dubal}, Pratik and {Krishnan}, Praveen and {Singh Koura}, Punit and {Xu}, Puxin and {He}, Qing and {Dong}, Qingxiao and {Srinivasan}, Ragavan and {Ganapathy}, Raj and {Calderer}, Ramon and {Silveira Cabral}, Ricardo and {Stojnic}, Robert and {Raileanu}, Roberta and {Maheswari}, Rohan and {Girdhar}, Rohit and {Patel}, Rohit and {Sauvestre}, Romain and {Polidoro}, Ronnie and {Sumbaly}, Roshan and {Taylor}, Ross and {Silva}, Ruan and {Hou}, Rui and {Wang}, Rui and {Hosseini}, Saghar and {Chennabasappa}, Sahana and {Singh}, Sanjay and {Bell}, Sean and {Kim}, Seohyun Sonia and {Edunov}, Sergey and {Nie}, Shaoliang and {Narang}, Sharan and {Raparthy}, Sharath and {Shen}, Sheng and {Wan}, Shengye and {Bhosale}, Shruti and {Zhang}, Shun and {Vandenhende}, Simon and {Batra}, Soumya and {Whitman}, Spencer and {Sootla}, Sten and {Collot}, Stephane and {Gururangan}, Suchin and {Borodinsky}, Sydney and {Herman}, Tamar and {Fowler}, Tara and {Sheasha}, Tarek and {Georgiou}, Thomas and {Scialom}, Thomas and {Speckbacher}, Tobias},
        title = "{The Llama 3 Herd of Models}",
      journal = {arXiv e-prints},
         year = 2024,
        month = jul,
          eid = {arXiv:2407.21783},
        pages = {arXiv:2407.21783},
          doi = {10.48550/arXiv.2407.21783},
archivePrefix = {arXiv},
       eprint = {2407.21783},
 primaryClass = {cs.AI},
       adsurl = {https://ui.adsabs.harvard.edu/abs/2024arXiv240721783G}
}

@article{gemini,
  title={Gemini: a family of highly capable multimodal models},
  author={Team, Gemini and Anil, Rohan and Borgeaud, Sebastian and Alayrac, Jean-Baptiste and Yu, Jiahui and Soricut, Radu and Schalkwyk, Johan and Dai, Andrew M and Hauth, Anja and Millican, Katie and others},
  journal={arXiv preprint arXiv:2312.11805},
  year={2023}
}

@article{gpt4,
  title={Gpt-4 technical report},
  author={Achiam, Josh and Adler, Steven and Agarwal, Sandhini and Ahmad, Lama and Akkaya, Ilge and Aleman, Florencia Leoni and Almeida, Diogo and Altenschmidt, Janko and Altman, Sam and Anadkat, Shyamal and others},
  journal={arXiv preprint arXiv:2303.08774},
  year={2023}
}

@ARTICLE{few_shot,
       author = {{Brown}, Tom B. and {Mann}, Benjamin and {Ryder}, Nick and {Subbiah}, Melanie and {Kaplan}, Jared and {Dhariwal}, Prafulla and {Neelakantan}, Arvind and {Shyam}, Pranav and {Sastry}, Girish and {Askell}, Amanda and {Agarwal}, Sandhini and {Herbert-Voss}, Ariel and {Krueger}, Gretchen and {Henighan}, Tom and {Child}, Rewon and {Ramesh}, Aditya and {Ziegler}, Daniel M. and {Wu}, Jeffrey and {Winter}, Clemens and {Hesse}, Christopher and {Chen}, Mark and {Sigler}, Eric and {Litwin}, Mateusz and {Gray}, Scott and {Chess}, Benjamin and {Clark}, Jack and {Berner}, Christopher and {McCandlish}, Sam and {Radford}, Alec and {Sutskever}, Ilya and {Amodei}, Dario},
        title = "{Language Models are Few-Shot Learners}",
      journal = {arXiv e-prints},
         year = 2020,
        month = may,
          eid = {arXiv:2005.14165},
        pages = {arXiv:2005.14165},
          doi = {10.48550/arXiv.2005.14165},
archivePrefix = {arXiv},
       eprint = {2005.14165},
 primaryClass = {cs.CL},
       adsurl = {https://ui.adsabs.harvard.edu/abs/2020arXiv200514165B}
}

@article{rest,
  title={Beyond human data: Scaling self-training for problem-solving with language models},
  author={Singh, Avi and Co-Reyes, John D and Agarwal, Rishabh and Anand, Ankesh and Patil, Piyush and Garcia, Xavier and Liu, Peter J and Harrison, James and Lee, Jaehoon and Xu, Kelvin and others},
  journal={arXiv preprint arXiv:2312.06585},
  year={2023}
}

@article{star,
  title={Star: Bootstrapping reasoning with reasoning, 2022},
  author={Zelikman, Eric and Wu, Yuhuai and Mu, Jesse and Goodman, Noah D},
  journal={URL https://arxiv. org/abs/2203.14465},
  volume={2203},
  year={2022}
}

@inproceedings{strubell2020energy,
  title={Energy and policy considerations for modern deep learning research},
  author={Strubell, Emma and Ganesh, Ananya and McCallum, Andrew},
  booktitle={Proceedings of the AAAI conference on artificial intelligence},
  volume={34},
  pages={13693--13696},
  year={2020}
}

@article{fernandez2025energy,
  title={Energy Considerations of Large Language Model Inference and Efficiency Optimizations},
  author={Fernandez, Jared and Na, Clara and Tiwari, Vashisth and Bisk, Yonatan and Luccioni, Sasha and Strubell, Emma},
  journal={arXiv preprint arXiv:2504.17674},
  year={2025}
}

@inproceedings{luccioni2024power,
  title={Power hungry processing: Watts driving the cost of ai deployment?},
  author={Luccioni, Sasha and Jernite, Yacine and Strubell, Emma},
  booktitle={Proceedings of the 2024 ACM conference on fairness, accountability, and transparency},
  pages={85--99},
  year={2024}
}

@article{wu2025unveiling,
  title={Unveiling environmental impacts of large language model serving: A functional unit view},
  author={Wu, Yanran and Hua, Inez and Ding, Yi},
  journal={arXiv preprint arXiv:2502.11256},
  year={2025}
}

@inproceedings{patel2024characterizing,
  title={Characterizing power management opportunities for llms in the cloud},
  author={Patel, Pratyush and Choukse, Esha and Zhang, Chaojie and Goiri, {\'I}{\~n}igo and Warrier, Brijesh and Mahalingam, Nithish and Bianchini, Ricardo},
  booktitle={Proceedings of the 29th ACM International Conference on Architectural Support for Programming Languages and Operating Systems, Volume 3},
  pages={207--222},
  year={2024}
}

@article{stojkovic2024towards,
  title={Towards greener llms: Bringing energy-efficiency to the forefront of llm inference},
  author={Stojkovic, Jovan and Choukse, Esha and Zhang, Chaojie and Goiri, Inigo and Torrellas, Josep},
  journal={arXiv preprint arXiv:2403.20306},
  year={2024}
}

@article{del2023skipdecode,
  title={Skipdecode: Autoregressive skip decoding with batching and caching for efficient llm inference},
  author={Del Corro, Luciano and Del Giorno, Allie and Agarwal, Sahaj and Yu, Bin and Awadallah, Ahmed and Mukherjee, Subhabrata},
  journal={arXiv preprint arXiv:2307.02628},
  year={2023}
}

@article{chen2023ee,
  title={Ee-llm: Large-scale training and inference of early-exit large language models with 3d parallelism},
  author={Chen, Yanxi and Pan, Xuchen and Li, Yaliang and Ding, Bolin and Zhou, Jingren},
  journal={arXiv preprint arXiv:2312.04916},
  year={2023}
}

@inproceedings{teerapittayanon2016branchynet,
  title={Branchynet: Fast inference via early exiting from deep neural networks},
  author={Teerapittayanon, Surat and McDanel, Bradley and Kung, Hsiang-Tsung},
  booktitle={2016 23rd international conference on pattern recognition (ICPR)},
  pages={2464--2469},
  year={2016},
  organization={IEEE}
}

@article{zhou2020bert,
  title={Bert loses patience: Fast and robust inference with early exit},
  author={Zhou, Wangchunshu and Xu, Canwen and Ge, Tao and McAuley, Julian and Xu, Ke and Wei, Furu},
  journal={Advances in Neural Information Processing Systems},
  volume={33},
  pages={18330--18341},
  year={2020}
}

@inproceedings{park2024large,
  title={Large Language Models are Students at Various Levels: Zero-shot Question Difficulty Estimation},
  author={Park, Jae-Woo and Park, Seong-Jin and Won, Hyun-Sik and Kim, Kang-Min},
  booktitle={Findings of the Association for Computational Linguistics: EMNLP 2024},
  pages={8157--8177},
  year={2024}
}

@inproceedings{dutulescu2024hard,
  title={How Hard Can This Question Be? An Exploratory Analysis of Features Assessing Question Difficulty Using LLMs},
  author={Dutulescu, Andreea and Ruseti, Stefan and Dascalu, Mihai and Mcnamara, Danielle},
  booktitle={Proceedings of the 17th International Conference on Educational Data Mining},
  pages={802--808},
  year={2024}
}

@inproceedings{xu2024adaption,
  title={Adaption-of-thought: Learning question difficulty improves large language models for reasoning},
  author={Xu, Mayi and Li, Yongqi and Sun, Ke and Qian, Tieyun},
  booktitle={Proceedings of the 2024 Conference on Empirical Methods in Natural Language Processing},
  pages={5468--5495},
  year={2024}
}

@inproceedings{talmor-etal-2019-commonsenseqa,
    title = "{C}ommonsense{QA}: A Question Answering Challenge Targeting Commonsense Knowledge",
    author = "Talmor, Alon  and
      Herzig, Jonathan  and
      Lourie, Nicholas  and
      Berant, Jonathan",
    editor = "Burstein, Jill  and
      Doran, Christy  and
      Solorio, Thamar",
    booktitle = "Proceedings of the 2019 Conference of the North {A}merican Chapter of the Association for Computational Linguistics: Human Language Technologies, Volume 1 (Long and Short Papers)",
    month = jun,
    year = "2019",
    address = "Minneapolis, Minnesota",
    publisher = "Association for Computational Linguistics",
    url = "https://aclanthology.org/N19-1421/",
    doi = "10.18653/v1/N19-1421",
    pages = "4149--4158"
}

@ARTICLE{2020arXiv200108361K,
       author = {{Kaplan}, Jared and {McCandlish}, Sam and {Henighan}, Tom and {Brown}, Tom B. and {Chess}, Benjamin and {Child}, Rewon and {Gray}, Scott and {Radford}, Alec and {Wu}, Jeffrey and {Amodei}, Dario},
        title = "{Scaling Laws for Neural Language Models}",
      journal = {arXiv e-prints},
         year = 2020,
        month = jan,
          eid = {arXiv:2001.08361},
        pages = {arXiv:2001.08361},
          doi = {10.48550/arXiv.2001.08361},
archivePrefix = {arXiv},
       eprint = {2001.08361},
 primaryClass = {cs.LG},
       adsurl = {https://ui.adsabs.harvard.edu/abs/2020arXiv200108361K}
}

@misc{gpt5,
  author = {OpenAI},
  title = {{OpenAI GPT5}},
  howpublished = {\url{https://openai.com/index/introducing-gpt-5/}},
  year = {2025}
}

\end{document}